\def\year{2021}\relax
\documentclass[letterpaper]{article} 
\usepackage{aaai21}  
\usepackage{times}  
\usepackage{helvet} 
\usepackage{courier}  
\usepackage[hyphens]{url}  
\usepackage{graphicx} 
\usepackage{natbib}  
\usepackage{caption} 
\usepackage{amsmath}
\usepackage{graphicx}
\usepackage{titlesec}
\usepackage{algorithm}  
\usepackage{algorithmic}  
\usepackage{subfigure}
\usepackage{rotating}
\title{Strategy and Benchmark for Converting Deep Q-Networks to Event-Driven Spiking Neural Networks}
\author{
    Weihao Tan, \textsuperscript{\rm 1}\thanks{Corresponding author}
    Devdhar Patel, \textsuperscript{\rm 1}
    Robert Kozma\textsuperscript{\rm 1}\textsuperscript{\rm 2}\\
}
\affiliations{
    \textsuperscript{\rm 1}Biologically Inspired Neural and Dynamical Systems Laboratory (BINDS), College of Computer and Information Sciences, 140 Governors Drive, University of Massachusetts Amherst, Amherst, MA 01003, USA\\
    
    \textsuperscript{\rm 2}Center for Large-Scale Integrated Optimization and Networks  (CLION) Department of Mathematical Sciences, 373 Dunn Hall, University of Memphis, Memphis, TN 38152, USA\\
    \{weihaotan, devdharpatel\}@cs.umass.edu,  rkozma55@gmail.com

}

\begin{document}

\maketitle

\begin{abstract}
Spiking neural networks  (SNNs) have great potential for energy-efficient implementation of Deep Neural Networks (DNNs) on dedicated neuromorphic hardware. Recent studies demonstrated competitive performance of SNNs compared with DNNs on image classification tasks, including CIFAR-10 and ImageNet data. The present work focuses on using SNNs in combination with deep reinforcement learning in ATARI games, which involves additional complexity as compared to image classification. We review the theory of converting DNNs to SNNs and extending the conversion to Deep Q-Networks (DQNs). We propose a robust representation of the firing rate to reduce the error during the conversion process. In addition, we introduce a new metric to evaluate the conversion process by comparing the decisions made by the DQN and SNN, respectively. We also analyze how the simulation time and parameter normalization influence the performance of converted SNNs. We achieve competitive scores on 17 top-performing Atari games. To the best of our knowledge, our work is the first to achieve state-of-the-art performance on multiple Atari games with SNNs. Our work serves as a benchmark for the conversion of DQNs to SNNs and paves the way for further research on solving reinforcement learning tasks with SNNs.
\end{abstract}

\section{Introduction}
Deep convolutional neural networks (CNNs) have made tremendous successes in image recognition in recent years \cite{lecun2015deep, he2016deep}. CNNs trained by deep reinforcement learning reached and even surpassed human-level performance in many RL tasks\cite{mnih2015human,vinyals2019grandmaster}. However, deep reinforcement learning combined with DNNs requires massive resources, which may not be available in certain practical problems. For example, in many real-life scenarios, the processing system is required to be power-efficient and to have low latency \cite{cheng2016wide}.

Spiking neural networks  (SNNs) provide an attractive solution to reduce latency and load of computation. As opposed to artificial neural networks  (ANNs) processing continuous activation values, SNNs use spikes to transmit information via many discrete events. This makes SNNs more power-efficient. SNNs can be implemented on dedicated hardware, such as IBM’s TrueNorth \cite{merolla2014million}, and Intel’s Loihi \cite{davies2018loihi}. They are reported to be 1000 times more energy-efficient than conventional chips. Thus, SNNs have the potential to answer the fast-growing demand toward AI as it finds its applications in many sectors of the society.

Theoretical analysis shows that SNNs are as computationally powerful as conventional neuronal models \cite{maass2004computational}, nevertheless, the use of SNNs is very limited at present. The main reason is that training multi-layer SNNs is a challenge, as the activation functions of SNNs are usually non-differentiable. Thus, SNNs cannot be trained with gradient-descent based methods, as opposed to ANNs. There are various alternative methods for training SNNs, such as using stochastic gradient descent \cite{o2016deep}, treating the membrane potentials as differentiable \cite{lee2016training}, event-driven random backpropagation \cite{neftci2017event}, and other approaches \cite{tavanaei2019bp, kheradpisheh2018stdp, mozafari2018combining}. These methods have a competitive performance on MNIST \cite{lecun1998gradient} with shallow networks compared with ANNs. However, none of them have a competitive performance on more realistic and complicated datasets, such as CIFAR-10 \cite{krizhevsky2012imagenet} and ImageNet \cite{russakovsky2015imagenet}. 
Recently,  \cite{wu2019direct} and \cite{lee2020enabling} used spike-based backpropagation to achieve more than 90\% accuracy on CIFAR-10, which is great progress for directly training SNNs.

To avoid the difficulty of training SNNs, at the same time taking advantage of the SNN processing, we follow an alternative approach of converting ANNs to SNNs proposed by \cite{perez2013mapping}. This method takes the parameters of a pre-trained ANN and maps them into an equivalent SNN with the same architecture \cite{diehl2015fast, rueckauer2017conversion, sengupta2019going}. As there are many high-performance training methods of ANNs, our goal to is directly convert state-of-the-art ANNs into SNNs without degradation in the performance. Recent work by \cite{sengupta2019going} demonstrated that this conversion is indeed possible. Namely, they showed that even very deep ANNs, like VGG16 and ResNet can be converted successfully to SNN with less than 1\% error on the CIFAR-10 and ImageNet dataset.

Following the successes of the ANN to SNN conversion method on image classification tasks, it is desirable to extend those results to solve deep reinforcement learning problems. However, using deep SNNs to solve reinforcement learning tasks proved to be very difficult. The problem can be summarized as follows. Due to the nature of the error used during optimization for well-trained ANN for image classification, the score of the correct class is always significantly higher than the other classes. Thus the error of the conversion process does not have a significant impact on the final performance. However, for DQNs, the output q-values of different actions are often very similar even for a well-trained network. This makes the network more susceptible to the error introduced during the conversion process. This is a key reason why converting DQNs to SNNs is a challenging task. 

In this paper, we propose SNNs to solve deep reinforcement learning problems. Our approach provides a more accurate approximation of the firing rates of spiking neurons as compared to the equivalent analog activation value proposed by \cite{rueckauer2017conversion}. First, we explore and explain the effect of parameter normalization described by \cite{rueckauer2017conversion}. Next, we propose a more robust representation of firing rate by using the membrane voltage at the end of the simulation time to reduce the error caused during the conversion process. Finally, we test the proposed method of converting DQNs to SNNs on 17 Atari games. All of them reach performances comparable with the original DQNs. This work explores the feasibility of employing SNNs in the deep reinforcement learning field. Successful implementations facilitate the use of SNNs in a wide range of machine learning tasks in the future. 

\section{Related Work}
In order to develop an efficient way to process data collected from event-driven sensors,  \cite{perez2013mapping} converts conventional frame-driven convolutional networks into event-driven spiking neural networks by mapping weights parameters from the convolutional network to SNN. However, this method requires tuning the parameters, which characterize the dynamics of the individual neurons. Furthermore, it suffers from considerable loss in accuracy. With the success of CNNs, researchers start to focus on using SNNs to solve computer vision problems, particularly in image classification and object recognition with the conversion method. \cite{cao2015spiking} is the first to report high accuracy on CIFAR-10 with SNNs converted from CNNs. It uses ReLU to guarantee all the activation values are non-negative and simple integrate-and-fire (IF) spiking neurons to eliminate the need for tuning extra parameters; the description of the IF neuron model is given in the next section. Their method had multiple simplifications; e.g., it requires to set all biases to zero after each training iteration, and it uses spatial linear subsampling instead of the more popular max-pooling. 

Based on \cite{cao2015spiking}, \cite{diehl2015fast} proposed a novel weight normalization method to achieve that the activation values of ANNs and firing rate of SNNs are in the same range. They reported 99.1\% test accuracy on MNIST.  \cite{rueckauer2017conversion} presented the comprehensive mathematical theory of the conversion method. They equated the approximation of the firing rate of spiking neurons to the equivalent analog activation value. After the spiking neuron generates a spike, their method subtracts the threshold of the neuron from the membrane potential, instead of resetting the membrane potential to zero. They observed that this approach can decrease the error. Furthermore, they proposed a more robust weight normalization based on the data-based normalization in  \cite{diehl2015fast} to alleviate the low firing rate in the higher layers. They also provided methods to convert many common operations in ANNs to their equivalent SNN versions, such as max-pooling,  softmax, batch-normalization and Inception-modules. With these methods, some deep neural network architectures, such as VGG-16 and Inception-v3 have been converted successfully to SNNs. They reported good performances on MNIST, CIFAR-10, and ImageNet. 

\cite{sengupta2019going} proposed a new method to adjust the threshold according to the input data, instead of doing weights normalization.
They successfully converted ResNets to SNNs. Their work achieves the state-of-the-art for conversion methods on CIFAR-10 and ImageNet. \cite{rueckauer2018conversion}
proposed the use of time-to-first-spike instead of firing rate to represent the activation values of ANN. Though this method greatly decreases the amount of computation, it introduces further errors. It reports to have a test accuracy loss of about 1\% on MNIST compared to the firing rate based conversion method, which will be scaled up in more complicated datasets and tasks.

Various studies aim at applying the conversion method for other tasks to take advantage of SNNs, such as event-based visual recognition  \cite{ruckauer2019closing} and natural language processing(NLP) \cite{diehl2016conversion}. \cite{patel2019improved} first attempted to use SNNs to do deep reinforcement learning task. It uses particle swarm optimization  (PSO) to search for the optimal weight scaling parameters. Though this method has a performance close to DQN using a shallow network and it has improved robustness to perturbations, it fails on deep DQN networks due to the large space of exploration. 

\section{Methods}
In this section, we first provide the mathematical framework for the conversion method, extending the work by \cite{rueckauer2017conversion}. Based on the developed theory, we propose two methods to reduce the error and increase the firing rate in the SNNs.

\subsection{Theory for converting ANNs to SNNs}
The main idea of the conversion method is to establish a relation between the firing rate of SNN and the activation value of ANN. With this relation, we can map the well-trained weights of ANNs to SNNs, so high-performance SNNs can be obtained directly, without the need for additional training. \cite{rueckauer2017conversion} reviews the state-of-art and proposes a complete mathematical theory of how to make an approximation of the firing rate of spiking neuron to the equivalent analog activation value. Their derivation had some over-simplifications, which can lead to errors in the conversion process. In this work, we rectified those inaccuracies and made the mathematical results more robust. The corrections introduced in the present work will be important only if the chosen threshold of the spiking neurons is not one. 

The spiking neuron we use is the IF neuron, which is one of the simplest spiking neuron models. The IF neuron simply integrates its input until the membrane potential exceeds the voltage threshold and a spike is generated. IF neuron does not have a decay mechanism, and we assume that there is no refractory period, which is more similar to the artificial neuron. 

For a SNN with L layers, let $W_l,b_l, l \in \{1,...,L\}$ denote the weight and bias of layer $l$. The input current of layer $l$ at time $t$, $z_l(t)$ can be computed as:
\begin{equation}
   z_l(t) = W_l \theta_{l-1}(t) + b_l
\end{equation}
where $\theta_{l-1}(t)$ is a matrix denoting whether the neurons in the layer $l-1$ generate spikes at time $t$:
\begin{equation}
    \theta_{l}(t) = \theta(V_l(t-1)+z_l(t)-V_{thr}),  \textrm{with} \  \theta(x)=\begin{cases} 1 \ x \ge 0 \\ 0 \ x < 0\end{cases}
\end{equation}
where $V_l(t-1)$ denotes the membrane potential of the spiking neurons in layer l at time $t-1$. $V_{thr}$ is the threshold of the spiking neuron. 

Here, our equation (1) is a little different from the one in \cite{rueckauer2017conversion}, whose equation is $z_l(t) = V_{thr}(W_l \theta_{l-1}(t) + b_l)$. If all the input currents are multiplied by the threshold, it is equal to setting the threshold to one. Then it results that whatever threshold we set, the input currents will change the corresponding multiple, which counters the effect of the threshold.

Normally, the typical IF neuron will reset its membrane potential to zero after it generates a spike. However, the message of the voltage surpassing the threshold is missing. \cite{rueckauer2017conversion} has proofed that it will introduce a new error and proposes a solution to eliminate it. When the membrane potential of a neuron exceeds the threshold, instead of resetting it to zero, subtracts the voltage of the threshold from the membrane potential. In this work, we also adopt this subtracting mechanism for all the IF neurons:
\begin{equation}
   V_l(t) = V_l(t-1) + z_l(t) - V_{thr}\theta_l(t)
\end{equation}

Equations (1) and (3) illustrate the main mechanism of IF neuron. Based on these two equations, we will attempt to establish the relationship between the firing rate of SNNs and the activation values of ANNs.

When $l > 0$, by accumulating equation (3) over the simulation time $t$, we can conclude:
\begin{equation}
   \sum_{t' = 1}^t V_l(t') = \sum_{t' = 1}^t V_l(t'-1) + \sum_{t' = 1}^t z_l(t') - \sum_{t' = 1}^t V_{thr}\theta_l(t')
\end{equation}
Inserting equation (1) into equation (4) and dividing the whole equation by $t$, we can conclude:

\begin{equation}
   \frac{V_l(t) -V_l(0)}{t} = \frac{\sum_{t' = 1}^t W_l \theta_{l-1}(t')-V_{thr}\theta_l(t') +  b_l}{t} 
\end{equation}
where $V_l(0) = 0$, $\frac{ \sum_{t' = 1}^t\theta_l(t')}{t}=r_l(t)$, which is the firing rate in layer $l$ at time $t$. To be simple, time step size is set to 1 unit, so $\frac{\sum_{t' = 1}^t 1}{t} = 1$. Equation(5) can be simplified as:
\begin{equation}
   \frac{V_l(t)}{t} = W_lr_{l-1}-V_{thr}r_l(t) + b_l
\end{equation}
\begin{equation}
   r_l(t) = \frac{W_lr_{l-1}(t) +b_l -\frac{V_l(t)}{t}}{V_{thr}}
\end{equation}
Let $\frac{V_l(t)}{t}$ be $\nabla V_l$. This yields:
\begin{equation}
   r_l(t) = \frac{W_lr_{l-1}(t) +b_l -\nabla V_l}{V_{thr}} 
\end{equation}
This equation is the core equation, which describes the relation of firing rate between adjacent layers. Similar with ANNs, the the firing rate in layer $l$ is mainly given by the sum of wights multiplied by the firing rate in the previous layer $l-1$ and bias. But the firing rate should also minus the approximation error caused by missing the information carried by the leftover membrane potential at the end of the simulation time. In addition, it is obvious that the firing rate is inversely proportional to the threshold. 

When $l = 0$, the layer $0$ is the input layer. In order to relate the firing rate of SNN and the activation values of ANN, we make them have the same input, which means  $z_0 = a_0$, where $a_0$ is the input value of ANN. Note that $a_0$ is scaled to $[0, 1]$ to have the same range as $z_0$, which can be done before or after training the ANN. It yields:
\begin{equation}
   V_0(t) -V_0(0) =  \sum_{t' = 1}^t z_0(t') - \sum_{t' = 1}^t V_{thr}\theta_0(t')
\end{equation}
\begin{equation}
   V_0(t) =  t a_0 - V_{thr}tr_0(t)
\end{equation}
\begin{equation}
   r_0(t) = \frac{a_0 -\nabla V_0}{V_{thr}}
\end{equation}
By repeating inserting expression (8) to the previous layer, starting with the first layer equation (11), we conclude:
\begin{equation}
    \begin{aligned}
       r_l(t) =&  \frac{\prod_{i=1}^lW_ia_0}{V_{thr^{l+1}}} + \frac{b_l}{V_{thr}} + \frac{W_lb_{l-1}}{V_{thr}^2}  +...+\frac{\prod_{i=2}^lW_ib_1}{V_{thr}^{l}} \\
       -& \frac{\nabla V_l}{V_{thr}} - \frac{W_l\nabla V_{l-1}}{V_{thr}^2} -...- \frac{\prod_{i=1}^lW_i\nabla V_0}{V_{thr}^{l+1}}
    \end{aligned}
\end{equation}
However, the final equation proposed by \cite{rueckauer2017conversion} is:
\begin{equation}
   r_l(t) = a_l - \nabla V_l - W_l\nabla V_{l-1} -...-\prod_{i=1}^lW_i\nabla V_0
\end{equation}
where $a_l$ is the activation value of ANN in the layer $l$.\\
As has been mentioned, in that work the threshold is multiplied unnecessarily. So if we set $V_{thr}$ set to 1 in equation (12), which is also what we do in the experiments, equations (12) and (13) will be very similar. This yields:
\begin{equation}
    \begin{aligned}
       r_l(t) =& \prod_{i=1}^lW_i a_0 + b_l + W_lb_{l-1} + ...+ \prod_{i=2}^lW_ib_1\\
       -& \nabla V_l - W_l\nabla V_{l-1} - ...- \prod_{i=1}^lW_i\nabla V_0
    \end{aligned}
\end{equation}
Though both equations have the same error part, the weight and bias part are different. \cite{rueckauer2017conversion} overlooks the activation functions: 
\begin{equation}
    \begin{aligned}
        a_l =& W_l(ReLu(W_{l-1}(...ReLu(W_1a_0 + b_1)) + b_{l-1})) +b_l \\
        \not=& W_l(W_{l-1}(...W_1a_0 + b_1) + b_{l-1}) +b_l\\
        \not=& \prod_{i=1}^lW_i a_0 + b_l + W_lb_{l-1} + ...+ \prod_{i=2}^lW_ib_1
    \end{aligned}
\end{equation}
So actually, there is no direct relationship between the firing rate of SNN and the activation value of ANN. The firing rate cannot be simply presented as the sum of activation value of ANN and some error terms  directly. But the firing rate is definitely an estimation of the activation value of ANN, though without activation functions. Let's back to equation (8). Though it doesn't apply ReLU explicitly, it does contain the ReLU mechanism inside. Because a spiking neuron can only generate one spike at one time step, $ 0\leq \sum_t\theta_l(t) \leq t$. And $r_l(t) = \sum_t\frac{\theta_l(t)}{t}$. The range of firing rate is $[0, 1]$. It guarantees that all the inputs to the next layer are greater or equal to 0, which has the same effect with ReLU.

The error term of equation (8), $\nabla V_{l}$, which is the membrane potential remaining in the neurons at the end of the simulation time, is dependent on the total input current of the previous layer, $W_lr_{l-1}(t)+b_l$. And it is worth noting that the membrane potential can be negative if the spiking neurons receive negative currents. So, there are four cases for equation (8).  
To be simple, let $R = \frac{W_lr_{l-1}(t) +b_l}{V_{thr}}, \nabla V = \frac{\nabla V_l}{V_{thr}}$.
\begin{itemize} 
\item \textbf{$R > 0, \nabla  V > 0$}. This is one of the most common cases. The spiking neuron receives positive currents and remains positive membrane potential, which is less than the threshold. $\nabla V = \frac{V_{l}(t)}{tV_{thr}}\in [0, \frac{1}{t}) \Rightarrow  R - \frac{1}{t}  < r_l(t) \leq R $
\item \textbf{$R < 0, \nabla  V > 0$}. This is the only impossible case. If the total current is negative, the leftover membrane potential cannot be positive.
\item \textbf{$R > 0, \nabla  V < 0$}. Though the total current is positive, before the end of the simulation time after the final spike, the spiking neuron starts to receive negative currents, which makes the final membrane potential negative. It is just like that the neuron makes an overdraft of spike. Different from ANNs, the firing rate can be different from the same firing rate as input due to the various sequence of input spikes. In this case, $r_l(t) > R$. 
\item \textbf{$R < 0, \nabla  V < 0$}. This case is also very common. One of the most common examples is that the spiking neuron continues to receive negative currents and generates no spikes during the simulation time. So the total input to the next layer is 0. This is just like the negative activation values of ANNs are reset to 0 by ReLU. However, due to the sequence of input current, spikes can be generated during the simulation time though the total current is negative. For example, at first, the spiking neuron continues to receive positive currents and generate a spike. Then it starts to receive negative currents until the end of the time. And the total input current is negative. So, in this case, the firing rate is unpredictable, $r_l(t) \in [0,1)$.
\end{itemize}

According to the discussion above, the error during the conversion process is caused by missing the part of the information carried by the leftover membrane potential at the end of the simulation time.
Contrast with the conclusion in \cite{rueckauer2017conversion} that the firing rate of a neuron is slightly lower than the corresponding activation value due to the reduction of the error caused by the membrane potential, actually, the firing rate doesn't have certain size relation with the corresponding activation value. Increasing the simulation time can alleviate the error. And we observe that the firing rate in the last layer is usually lower than the corresponding activation values. This can be partly explained that because \textbf{$R > 0, \nabla  V > 0$} is the most common case, the general trend of the firing rate is decreasing. The error accumulated layer by layer and the final firing rate is therefore lower than the activation value.

\subsection{Methods to Alleviate the Error}
In this section, We introduce a set of methods to alleviate the error during the conversion process.

\subsubsection{Parameter Normalization}
The first data-based parameter normalization is put forward by \cite{diehl2015fast}, improved by \cite{rueckauer2017conversion}. The purpose of this operation is to scale the activation value to the same range as the firing rate. Then the weights and biases can be mapped into SNNs directly. It uses the training set to collect all the maximum activation values of every ReLU layer. By dividing by these maximum values, all the ReLU activation values are scaled to [0, 1].
\begin{equation}
    w_l^n = \frac{\lambda_{l-1}}{\lambda_l}w_l \quad b_l^n = \frac{b_l}{\lambda_l}
\end{equation}
where $w_l^n$ and $b_l^n$ are the normalized weight and bias. $w_l^n$ and $b_l^n$ are the original weight and bias. $\lambda_l$ is the maximum ReLU values of layer l. 

Based on the observation that there are a few activation values considerably greater than the others, which will make most of the activation values considerably less than 1, \cite{rueckauer2017conversion} improves the normalization method and proposes a more robust one. Instead of using the maximum activation value of the layer to do the normalization, they use the p-th percentile activation value of the layer. P is a hyperparameter, which usually has a good performance in the range of [99.0, 99.999]. This method intends to improve the firing rate, thereby decreasing the latency until the spikes reach the deeper layers.

In the deep reinforcement learning tasks, we find the performance of the converted SNNs are very sensitive to the hyperparameter, p(percentile). It not only decreases the latency among the layers but also greatly improves the performance of the converted SNNs from DQNs. This is because by tuning the percentile, the firing rate increases, and the error caused by the leftover membrane potential is partly compensated. Thus, the accuracy gap between the original DQNs and the converted SNNs is alleviated. In this work, we make an assumption that all the spiking neurons have the same threshold. Though there are works decreasing the error by tuning the threshold in different layers separately \cite{sengupta2019going}. It actually has the same effect theoretically with changing the percentile of the parameter normalization.  

\subsubsection{Robust Firing Rate}
According to equation (7), the error is inversely proportional to the simulation time. The longer simulation time we set, the smaller influence will the firing rate be affected. However, longer simulation means less power-efficient. So it is a trade-off between accuracy and efficiency. Normally the total simulation time is set to 100-1000 time steps. The firing rate is actually discrete with limited simulation time.
As has been mentioned above, the q-values among different actions are very close. This makes the spiking neurons in the final layer of SNNs sometimes receive the same number of spikes, resulting in the same firing rate. This situation is particularly frequent when the simulation time is not long enough. In order to distinguish the real optimal action among spiking neurons with the same firing rate. We introduce the leftover membrane potential to solve this issue. The intuition of this method is that 
the neurons with higher membrane potential are more likely to generate the next spike. So when there are several neurons with the same firing rate, we should choose the neuron with the highest membrane potential. Theoretically, adding membrane potential also has its own benefit. According to equation (13), $\nabla V_l = \frac{V_l(t)}{t}$ is the only part of error that we can get directly. $V_l(t)$ is exactly the membrane potential at the end of the simulation time. Combining with equation (7),  adding the membrane potential can eliminate the error in the last layer, which makes the firing rate of SNNs closer to the activation values of ANN. 
\begin{equation}
    f_{last}(t) = r_{last}(t) + \frac{V_{last}(t)}{tV_{thr}}
\end{equation}
where $r_{last}(t)$ and $V_{last}(t)$ are the firing rate and membrane potential of spiking neurons in the last layer at time t. $f_{last}(t)$ is the proposed more robust representation of the firing rate of spiking neurons in the last layer at time $t$. The robust firing rate not only increases the accuracy but also decreases the latency. It makes SNNs take less simulation time to distinguish the size relationships among the different actions. 

\begin{figure*}[htb]
  \centering
  \subfigure[The average score and conversion rate of converted SNNs with different simulation time on different four games. The percentile is set to 99.9 (Breakout), 99.99 (James Bond), 99.9 (Star Gunner) and 99.9 (Video Pinball).]{
      \includegraphics[scale=0.5]{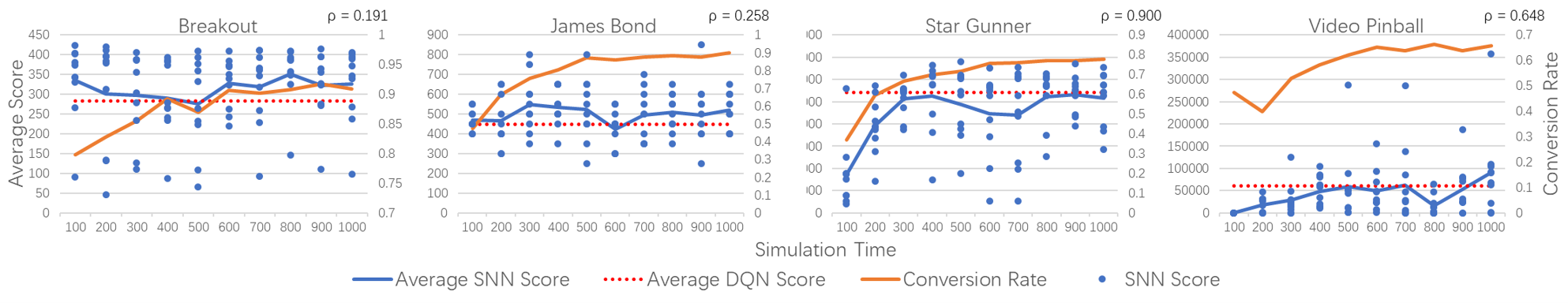}
      \label{fig:Time}
  }
  \subfigure[The average score and conversion rate of converted SNNs with different percentile to do the parameter normalization on different four games. The simulation time is set to 500 time steps.]{
      \includegraphics[scale=0.5]{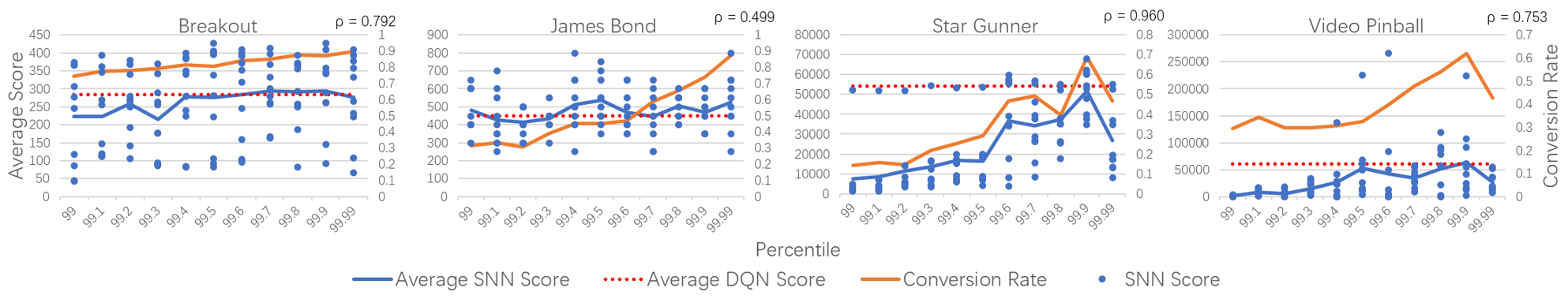}
      \label{fig:Percentile}
  }
  \caption{ Four games, Breakout, James Bond, Star Gunner and Video Pinball, are tested with different simulation time and percentile. Each game is played 10 times with different hyperparameters (simulation time and percentile). The blue and orange lines are the average score and conversion rate of converted SNNs. The blue dots are the scores obtained by SNNs in 10 times. The red dot line is the average score of DQN over 10 times. The Pearson correlation coefficient of average SNN score and conversion rate is shown at the top right corner. }
  \label{fig:TimeAndPercentile}
\end{figure*}

\section{Results}
In this section, we use Atari games to evaluate our approach. Code used for experiments can be found at :
https://github.com/WeihaoTan/bindsnet-1.

\subsubsection{Environment}
The arcade learning environment (ALE) \cite{bellemare2013arcade} is a platform that provides an interface to hundreds of Atari 2600 game environments. It is designed to evaluate the development of general, domain-independent AI technology. The agent obtains the situation of the environment through image frames of the game, which is revised to 84X84 pixels, interacts with the environment with at most 18 possible actions, and receives feedback in the form of the change in the game score. Since \cite{mnih2015human} shows human-level control of the agent on Atari games, Atari games have become a basic benchmark for deep reinforcement learning and ALE is a very popular platform for Atari games.

Without dedicated SNNs hardware, simulating firing-rate-based SNNs is a computationally demanding task, especially for reinforcement learning tasks. We choose to use BindsNET\cite{hazan2018bindsnet} to simulate SNNs. BindsNET is the first and only open-source library used for simulating SNNs on PyTorch Platform. It takes advantage of GPUs, greatly accelerating the process of testing SNNs. All the results are run on a single NVIDIA Titan-X GPU.

\subsubsection{Networks}
To obtain the converted SNNs, we need to train DQNs first. The DQN we use is the naive DQN proposed by \cite{mnih2015human} with 3 convolution layers and 2 fully connected layers. The first convolution layer has 32 8X8 filters with stride 4. The second convolution layer has 64 4X4 with stride 2. The third convolution layer has 64 3X3 filters with stride 1. The fully connected layer has  512 neurons. The number of neurons in the final layer depends on the number of valid actions of the tested games. We use ReLU as the activation function, which is applied to every layer except the final layer. All the games are trained with this architecture. The hyperparameters we use are also the same with \cite{mnih2015human}.

\subsubsection{Conversion}
After obtaining well-trained DQNs, we convert them to the corresponding SNNs. Firstly, we need to use IF neurons to build SNNs with the same architecture as DQNs. Before we do the parameter normalization, we need to collect the game frames to calculate the p-th maximum activation values of every layer. For each game, we collect at most 15000 frames to do the parameter normalization, then we map these parameters to the corresponding spiking neurons. Finally, we obtain the converted SNNs. 

\subsubsection{Metric}
For Atari games, the score obtained in the game is the main metric to evaluate the performance of the method. However, sometimes, the average score cannot represent the performance well due to the variance if the games are not run for a sufficient number of times. To test the effect of the conversion process more efficiently and to show the result more directly, we introduce a new metric, conversion rate. It is inspired by the accuracy in image classification, and it is represented as follow:
\begin{equation}
    CR = \frac{E}{NA}
\end{equation}
where $CR$ is the conversion rate, $E$ is the number of actions that the SNNs choose the same with DQNs in one game. $NA$ is the total number of actions that the agent takes in the game. The conversion rate is very similar to the accuracy in the image classification task. The actions chosen by DQNs can be viewed as proxies of the correct labels. The conversion rate describes the similarity between the behavior of the SNN agent and the DQN agent. The conversion rate can be obtained by feeding the frames of the game to both the original DQN and the converted SNN simultaneously, but the agent only takes the action given by the SNN. Then we can get the conversion rate and the game score of SNN at the same time.

\subsubsection{Testing}
To test the performance of the converted SNNs, each episode of the game is evaluated up to 5 minutes (18,000 frames) per run. To test the robustness of the agent in different situations, the agent starts by random times (at most 30 times) of no-op actions. A greedy policy is applied to minimize the possibility of overfitting during the evaluation, where $\epsilon = 0.05$. If not mentioned, the firing rate we use is the robust firing rate.

\begin{table*}[htb]
\centering
\caption{Comparison of performance of different games obtained by DQN agents and converted SNN agents. Each game is run for 50 times. The simulation time is set to 500 time steps.}
\label{table:games}
\begin{tabular}{ccccc}
\hline
Game           & DQN Score ($\pm$ std)  & SNN Score ($\pm$ std)     & Normalized SNN (\% DQN) & P-Value\\ \hline
Krull          & 3389$\pm$667.3        & 4128.2$\pm$900.4     & 121.8\%   & 0.0002             \\
Tutankham      & 147.1$\pm$51.8        & 157.1$\pm$37.7       & 106.8\%     &0.770           \\
Video Pinball  & 69544.1$\pm$98993.2   & 74012.1$\pm$129840.3 & 106.4\%    &0.847            \\
Beam Rider     & 8703.6$\pm$3648.1     & 9189.3$\pm$3650.4    & 105.6\%     &0.522           \\
Name This Game & 8163.5$\pm$1394.7     & 8448$\pm$1457        & 103.5\%     &0.596           \\
Jamesbond      & 504.4$\pm$127.7       & 521$\pm$132.5        & 103.3\%     & 0.823           \\
SpaceInvaders  & 2193.4$\pm$927.8      & 2256.8$\pm$890.4     & 102.9\%      &0.932          \\
RoadRunner     & 41053.6$\pm$6534.9    & 41588$\pm$5071.2     & 101.3\%      & 0.820          \\
Atlantis       & 176044.1$\pm$170037.1 & 177034
$\pm$182105.7
    & 100.6\%      &0.760           \\ 
Pong           & 19.9$\pm$1.2          & 19.8$\pm$1.3         & 99.7\%           &0.448      \\
Kangaroo       & 9874.2$\pm$2573.2     & 9760$\pm$2499.3      & 98.8\%       &0.967          \\
Crazy Climber  & 109885.9$\pm$21633.9  & 106416$\pm$25946.7   & 96.8\%         &0.671        \\
Tennis         & 14$\pm$3.5            & 13.3$\pm$3.2         & 95.2\%          &0.280       \\
Star Gunner    & 57783.4$\pm$8202      & 54692$\pm$9963.3     & 94.7\%       & 0.038          \\
Gopher         & 7076.1$\pm$3570.9     & 6691.2$\pm$3839.2    & 94.6\%       &0.823          \\
Boxing         & 80$\pm$10.9           & 75.5$\pm$9.4         & 94.4\%       &0.356          \\
Breakout       & 311.9$\pm$101.7       & 286.7$\pm$117.2      & 91.9\%      &0.210           \\ \hline
\end{tabular}
\end{table*}

\subsection{The Effect of Simulation Time}
Figure \ref{fig:Time} shows the performance of SNNs with different simulation time on four games. If the simulation time is set to 100, the same frame will be fed into SNN as input for 100 times. The percentiles we choose are the ones with the best performance. From the figure, we can conclude that as the simulation time increases, the conversion rates of all the four games also increase, which is consistent with the conclusion in the Methods section that the error caused by the leftover membrane potential can be alleviated by the longer simulation time, but the effect of diminishing. It is a trade-off between the low error and the power-efficiency. 

For the games that not very sensitive to the simulation time, like Breakout and James Bond, even with very short simulation time like 100 time steps, the SNN agents can still have good performance with low conversion rates. These SNN models are more robust. The error always influences the size relationships among the actions with close q-values.
So the wrong actions chosen by the SNN agent can still be the sub-optimal action. The agent can still have good performance with the low conversion rate. For the games like Star Gunner and Video Pinball, these games are more sensitive to the simulation time, longer simulation time results in a larger conversion rate and then results in better performance.

\subsection{The Effect of Percentile}
Figure \ref{fig:Percentile} shows the performance of SNNs with different percentile to do the parameter normalization on four games. The percentile is a very sensitive hyperparameter. Numerical studies indicate that we are unlikely to get good performance when the percentile is less than 99. Thus, we tested the effect of the percentile starting from 99. For  all four games, the percentile greatly influences the conversion rate and the conversion rate affects the final score. For games like Breakout and James Bond, the conversion rate continuously improves as the percentile increases. Setting the percentile close to 100 can have good performance.  For the other games like Star Gunner and Video Pinball, there is a maximum point of performance as the percentile increases. In these two games, percentile 99.9 gives the maximum value. Once exceeding this point, the performance of these games will drop dramatically. This result shows that tuning the percentile value can greatly improve the performance of the agent on these games.

\subsection{The Performance of Converted SNNs on Different Games}
Table \ref{table:games} shows the performance of our method on multiple Atari games. Due to the limitation of resources, we choose 17 games to test the method. These are the games on which DQNs have very high performances. The total simulation time is 500 time steps. We choose the best percentile in [99.9, 99.99]. Every game is played 50 times. The average scores of DQNs and converted SNNs are shown in the table. The P-Values of DQNs and SNNs obtained by T-Test are also shown. We conclude that the scores of all SNNs are close to the scores of DQNs. And almost all the games, except Kull and Star Gunner, do not have a significant difference between DQN and SNN. This shows that the SNNs converted by our method have comparable performance to the original DQNs.

\section{Conclusions and Future Work}

This work addressed the issue of solving deep reinforcement learning problems using SNNs by converting the parameters of DQNs to SNNs. 
\begin{itemize}
\item
We propose a robust conversion method, which reduces the error caused during the conversion process. We tested the proposed method on 17 Atari games and showed that SNNs can reach performances comparable with the original DQNs. 
\item
It is important to point out that our approach provides a more accurate approximation of the firing rates of spiking neurons as compared to the equivalent analog activation value proposed by \cite{rueckauer2017conversion}.  Our work corrects some previous inaccuracies of the conversion method and further develops the theory. 
\item
To our best knowledge, our work is the first one to achieve high performance on multiple Atari games with SNNs. It paves the way to run reinforcement learning algorithms on real-time systems with dedicated hardware efficiently. 
\item
We do not have a very high conversion rate for a few games. There are some actions taken by the SNNs agent, which are not the optimal actions, which decreases the robustness of SNNs in those games. The reason for these issues is the objective of ongoing and future studies
\end{itemize}

\section{Acknowledgments}
This work has been supported in part by Defense Advanced Research Project Agency, USA Grant, DARPA/MTO HR0011-16-l-0006.

\bibliography{main}

\title{Supplymentary Material}
\subsection{Supplymentary Material}
Table \ref{table:full games} shows the detailed performance of our method on 17 Atari games. The total simulation time is 500 time steps. Every game is played 50 times. We show the SNN score and conversion rate with two different percentiles, 99.9 and 99.99, which are more likely to have good performance. The SNN Score is obtained with robust firing rate. Meanwhile we also show the conversion rate of SNN using normal firing rate. From the table, we can conclude that the robust firing rate can significantly improve the conversion rate compared to the normal firing rate. And SNNs can have comparable performance compared to DQNs by tuning the percentile. 

\begin{sidewaystable}[htb]
\centering
\caption{Comparison of games performance obtained by DQN agents and converted SNN agents}
\label{table:full games}
\resizebox{650pt}{100pt}{
\begin{tabular}{cccccccc}
\hline
& \multicolumn{1}{c|}{} & \multicolumn{3}{c|}{Percentile99.9}& \multicolumn{3}{c}{Percentile99.99} \\ \hline
\multicolumn{1}{l}{} & \multicolumn{1}{l|}{} & \multicolumn{1}{c|}{}   & \multicolumn{2}{c|}{Conversion Rate(\%)}    & \multicolumn{1}{l|}{}   & \multicolumn{2}{c}{Conversion Rate(\%)}                        \\ \hline
Game                 & ANN Score             & SNN Score               & Normal Firing Rate & Robust Firing Rate & SNN Score               & Normal Firing Rate & \multicolumn{1}{l}{Robust Firing Rate} \\ \hline

Atlantis       & 176044.1$\pm$170037.1 & 130342.0 $\pm$ 163829.1 & 80.7 $\pm$ 3.2\ & 89.8 $\pm$ 1.4\       & 177034 $\pm$ 182105.7 & 82.7 $\pm$ 2.8\  & 94.0 $\pm$ 1.2\        \\
Beam Rider     & 8703.6$\pm$3648.1     & 8174.2 $\pm$ 3457.9     & 67.0 $\pm$ 1.1\ & 73.9 $\pm$ 1.0\       & 9189.3 $\pm$ 3650.4     & 77.7 $\pm$ 0.9\  & 90.6 $\pm$ 0.5\                           \\
Boxing         & 80$\pm$10.9           & 78.0 $\pm$ 20.4         & 55.6 $\pm$ 9.8\ & 57.2 $\pm$ 10.0\      & 75.5 $\pm$ 9.4          & 76.6 $\pm$ 4.3\  & 85.8 $\pm$ 1.9\                           \\
Breakout       & 311.9$\pm$101.7       & 286.7 $\pm$ 117.2       & 87.0 $\pm$ 2.9\ & 87.0 $\pm$ 1.8\       & 268.2 $\pm$ 128.9       & 89.8 $\pm$ 2.8\  & 82.9 $\pm$ 3.6\                           \\
Crazy Climber  & 109885.9$\pm$21633.9  & 92566.0 $\pm$ 25259.3   & 42.6 $\pm$ 3.1\ & 61.4 $\pm$ 7.3\       & 106416.0 $\pm$ 25946.7  & 45.3 $\pm$ 3.1\  & 82.5 $\pm$ 3.1\                           \\
Gopher         & 7076.1$\pm$3570.9     & 5285.6 $\pm$ 2813.5     & 63.0 $\pm$ 3.8\ & 78.7 $\pm$ 2.0\       & 6691.2 $\pm$ 3839.2     & 72.1 $\pm$ 4.1\  & 92.5 $\pm$ 1.0\                           \\
Jamesbond      & 504.4$\pm$127.7       & 493.0 $\pm$ 122.9       & 67.4 $\pm$ 2.6\ & 74.5 $\pm$ 2.6\       & 521.0 $\pm$ 132.5       & 73.6 $\pm$ 1.5\  & 83.5 $\pm$ 1.3\                           \\
Kangaroo       & 9874.2$\pm$2573.2     & 9358.0 $\pm$ 2408.5     & 58.4 $\pm$ 1.5\ & 70.2 $\pm$ 1.2\       & 9760.0 $\pm$ 2499.3     & 65.7 $\pm$ 1.7\  & 87.3 $\pm$ 0.7\                           \\
Krull          & 3389$\pm$667.3        & 4128.2 $\pm$ 900.4      & 59.6 $\pm$ 3.0\ & 66.2 $\pm$ 2.6\       & 3697.2 $\pm$ 758.4      & 69.5 $\pm$ 2.2\  & 80.9 $\pm$ 1.9\                           \\
Name This Game & 8163.5$\pm$1394.7     & 8448.0 $\pm$ 1457.0     & 53.1 $\pm$ 1.4\ & 74.1 $\pm$ 1.3\       & 8488.0 $\pm$ 1677.9     & 56.0 $\pm$ 1.6\  & 71.6 $\pm$ 1.4\                           \\
Pong           & 19.9$\pm$1.2          & 19.4 $\pm$ 1.5          & 59.9 $\pm$ 1.7\ & 63.9 $\pm$ 1.7\       & 19.8 $\pm$ 1.3          & 72.0 $\pm$ 1.8\  & 85.8 $\pm$ 1.2\                           \\
RoadRunner     & 41053.6$\pm$6534.9    & 40344.0 $\pm$ 7080.2    & 56.0 $\pm$ 6.2\ & 71.6 $\pm$ 3.3\       & 41588.0 $\pm$ 5071.2    & 68.3 $\pm$ 2.4\  & 75.2 $\pm$ 3.3\                           \\
SpaceInvaders  & 2193.4$\pm$927.8      & 2133.0 $\pm$ 822.9      & 70.2 $\pm$ 1.8\ & 76.8 $\pm$ 1.6\       & 2256.8 $\pm$ 890.4      & 70.1 $\pm$ 1.8\  & 83.0 $\pm$ 1.6\                           \\
Star Gunner    & 57783.4$\pm$8202      & 54692.0 $\pm$ 9963.3    & 61.6 $\pm$ 1.5\ & 71.6 $\pm$ 1.4\       & 30580.0 $\pm$ 15472.6   & 44.2 $\pm$ 2.2\  & 42.7 $\pm$ 2.1\                           \\
Tennis         & 14$\pm$3.5            & 12.2 $\pm$ 4.3          & 59.4 $\pm$ 1.5\ & 68.0 $\pm$ 1.0\       & 13.3 $\pm$ 3.2          & 64.8 $\pm$ 1.7\  & 72.2 $\pm$ 1.3\                           \\
Tutankham      & 147.1$\pm$51.8        & 152.0 $\pm$ 44.3        & 88.9 $\pm$ 6.5\ & 93.4 $\pm$ 4.7\       & 157.1 $\pm$ 37.7        & 86.2 $\pm$ 8.6\  & 97.9 $\pm$ 1.2\                           \\
Video Pinball  & 69544.1$\pm$98993.2   & 74012.1 $\pm$ 129840.3  & 53.1 $\pm$ 9.3\ & 61.0 $\pm$ 9.9\       & 8.0 $\pm$ 39.6          & 59.6 $\pm$ 14.2\ & 42.7 $\pm$ 14.2\                          \\ \hline
\end{tabular}}
\end{sidewaystable}

\end{document}